\documentclass{article}

\usepackage[preprint]{neurips_2026}

\usepackage[utf8]{inputenc}
\usepackage[T1]{fontenc}
\usepackage{hyperref}
\usepackage{url}
\usepackage{booktabs}
\usepackage{amsfonts}
\usepackage{nicefrac}
\usepackage{microtype}
\usepackage{xcolor}

\usepackage{graphicx}
\usepackage{amsmath, amssymb}
\usepackage{tikz}
\usetikzlibrary{arrows.meta, positioning}
\usepackage{caption}
\usepackage{wrapfig}

\title{LLMs as Implicit Imputers:\\Uncertainty Should Scale with Missing Information}

\author{%
    Stef van Buuren \\
  1.\ TNO - Netherlands Organization for Applied Scientific Research, Leiden\\
  2.\ Dept.\ of Methodology and Statistics, University of Utrecht\\
  \texttt{stef.vanbuuren@tno.nl} \\
}

\begin{document}

\maketitle

\begin{abstract}
Large language models (LLMs) are increasingly deployed in settings where the available context is incomplete or degraded. We argue that an LLM generating answers under incomplete context can be viewed as an implicit imputer, and evaluated against a criterion from the multiple imputation (MI) literature: uncertainty should scale with the amount of missing information. We assess this criterion on SQuAD, using a controlled framework in which context availability is varied across five levels. We evaluate two answer-level uncertainty measures that can be estimated from repeated sampling: sampling-based confidence (empirical mode frequency) and response entropy. Confidence fails to reflect increasing missingness: it remains high even as accuracy collapses. Entropy, by contrast, increases with context removal, consistent with the MI analogy, and explains substantially more variance in accuracy than confidence across all evidence levels (quadratic $R^2$ gap up to 0.057). We further introduce a black-box diagnostic $\rho_R(\alpha)$ that estimates the proportion of baseline uncertainty resolved by context level $\alpha$, requiring only repeated sampling with and without context. These results suggest that entropy is a more responsive black-box uncertainty measure than confidence under incomplete context.
\end{abstract}

\section{Introduction}

Large language models (LLMs) are increasingly deployed as predictive systems in settings where relevant information is incomplete, truncated, or missing. In question answering, supporting context may be partially available; in retrieval-augmented generation, retrieved passages may be absent, noisy, or irrelevant. Despite operating under such degraded inputs, LLMs typically produce a single answer with high apparent certainty. This raises a fundamental question:

{\leftskip=1em\relax
\noindent Do LLM uncertainty estimates appropriately reflect the amount of missing information?
\par}

This question connects to a broader concern in modern AI. Recent work in Bayesian deep learning argues that predictive systems should represent and propagate uncertainty in a manner consistent with the available information: uncertainty should increase as inputs become less informative, reflecting posterior uncertainty about unobserved quantities \citep{papamarkou2024}. However, current large-scale models are typically evaluated under fixed inputs, leaving open whether their uncertainty estimates respond appropriately when information is systematically degraded.

In this paper, we operationalize this requirement in the setting of LLMs under incomplete context. Our approach is inspired by multiple imputation (MI) \citep{rubin1987}, a principled framework for inference with missing data. In MI, uncertainty arises from unobserved variables and is quantified through variability across multiple completed datasets. We view repeated sampling from an LLM under incomplete context as generating implicit completions of missing information, inducing a distribution over possible answers. This perspective yields a minimal, black-box criterion: a \emph{proper} predictive system should exhibit uncertainty that increases with the fraction of missing information.

We evaluate two answer-level uncertainty measures that can be estimated from repeated sampling without access to model internals: sampling-based confidence (the empirical mode frequency) and response entropy. While both are derived from the same sample distribution, they capture different aspects of predictive behavior. Entropy reflects dispersion across competing answers and serves as a proxy for predictive variability induced by missing information. Confidence, by contrast, measures modal concentration and can remain high even when the underlying distribution is diffuse.

Empirically, we show that response entropy increases monotonically as context is removed and closely tracks the resulting loss in accuracy. Confidence does not: it remains high even as predictive performance collapses. From the perspective of uncertainty propagation, entropy behaves in accordance with the Bayesian requirement that uncertainty should scale with missing information; confidence systematically fails this requirement.

\textbf{Contributions:}
\begin{itemize}
\item A principled evaluation framework for LLM uncertainty under partial observability, grounded in the requirement that uncertainty should increase as available information decreases.
\item An analogy between repeated LLM sampling and multiple imputation, motivating entropy as a black-box proxy for predictive uncertainty induced by missing information.
\item A controlled experimental setup that varies evidence completeness and isolates its effect on accuracy, entropy, confidence, and the resolution ratio $\rho_R(\alpha)$.
\item Empirical evidence that response entropy tracks missing-information uncertainty more faithfully than sampling-based confidence across evidence conditions, including a quantitative advantage in explained variance (quadratic $R^2$ gap up to 0.057).
\item A heuristic diagnostic, the \emph{Resolution Ratio} $\rho_R(\alpha)$, measuring the fraction of baseline uncertainty resolved by context level $\alpha$, estimable from black-box sampling alone.
\end{itemize}

\section{Related Work}

\textbf{LLM calibration and confidence elicitation.} A substantial body of work studies whether LLM confidence matches empirical accuracy under fixed inputs. \citet{guo2017} show that modern neural networks are systematically overconfident, a finding that extends to LLMs \citep{kadavath2022, xiong2024}. Related work examines verbalized confidence and token probabilities as uncertainty proxies \citep{kadavath2022, xiong2024}, and studies sample consistency from repeated generations \citep{taubenfeld2025}. We differ in asking whether uncertainty estimates \emph{respond} when the informativeness of the input is varied --- a dynamic rather than static calibration criterion.

\textbf{Uncertainty estimation from sampling.} \citet{lakshminarayanan2017} propose deep ensembles as a principled approach to predictive uncertainty. In the LLM setting, \citet{kuhn2023} introduce semantic entropy, which aggregates response diversity across semantically equivalent answers using an entailment classifier. Our entropy measure uses exact string matching over short factual answers, avoiding the entailment dependency and preserving the black-box property.

\textbf{Multiple imputation.} \citet{rubin1987} established the theoretical foundations of MI, including the variance decomposition and the concept of proper imputation. \citet{bartlett2015} extend this framework to substantive-model-compatible imputation, and \citet{little2020} provide a comprehensive treatment of missing data methods.

To our knowledge, the connection between MI and LLM uncertainty estimation has not been made explicitly. Rubin's combining rules summarize uncertainty through between-imputation variance for numerical estimands. In the present setting, however, model outputs are categorical, and uncertainty is expressed through dispersion over a set of possible answers. We therefore work directly with response entropy, which measures variability over the full predictive distribution and serves as a natural analogue of between-imputation uncertainty for categorical responses.

\section{Methods}

\subsection{Problem Formulation}

Let $q$ denote the question, $C$ the available context, and $y$ the ground-truth answer. We consider a sequence of evidence conditions $C_0 \prec C_1 \prec \dots \prec C_K$ representing increasing completeness. Given an input $(q, C)$, we obtain $m$ stochastic samples from a large language model:
\[
Y^{(1)}, \dots, Y^{(m)} \sim p_\theta(\cdot \mid q, C).
\]
Let $\tilde{Y}^{(l)}$ and $\tilde{y}$ denote normalized forms of the sampled responses and ground truth. Answers were considered correct if they matched the ground-truth entity after normalization, allowing for minor lexical variation (e.g., omission of articles or modifiers), but not for semantic generalization or substitution. We define correctness indicators
\[
Z^{(l)} = \mathbf{1}(\tilde{Y}^{(l)} = \tilde{y}).
\]
Accuracy is estimated as
\[
\mathrm{Acc} = \frac{1}{m} \sum_{l=1}^m Z^{(l)}.
\]
Our target quantity is
\[
p(q, C) = \mathbb{P}(Z = 1 \mid q, C),
\]
where $Z$ denotes the correctness indicator of a generic sampled response.

Each sampled response can be viewed as an implicit completion of missing information, and $p(q, C)$ quantifies the probability that such a completion yields the correct answer.

\subsection{Uncertainty Measures}

LLMs do not natively expose uncertainty at the level of complete answers. Although some APIs provide token-level log-probabilities, these quantify the model's distribution over next tokens. Such token-level probabilities do not capture the probability mass assigned to alternative valid answers. We therefore consider two uncertainty measures that can be estimated from repeated sampling alone, requiring no access to model internals and applicable to any LLM API.

Let $p_k$ denote the empirical frequency of normalized answer $k$ among the $m$ sampled responses. We characterize uncertainty using two complementary quantities.

\paragraph{Sampling-based confidence:} $c = \max_k p_k$, the empirical probability of the most frequent answer. This captures the degree of agreement among repeated samples and serves as a proxy for certainty.

\paragraph{Response entropy:} $H = -\sum_k p_k \log p_k$, which quantifies dispersion across distinct normalized answers. Higher entropy indicates that responses are spread across many different answers.

Both measures summarize the same distribution of sampled answers (confidence reflecting concentration, entropy reflecting spread) and are monotonically related in the two-answer case. With more diverse response distributions, their behavior can diverge.

\phantomsection\label{sec:rr}%
\paragraph{Resolution Ratio $\rho_R$:} Let $H(C)$ denote the response entropy under context $C$ and $H_0 = H(\emptyset)$ the entropy under no context. The \emph{Resolution Ratio} $\rho_R(C)$ is the fraction of baseline uncertainty resolved by context $C$:
\[
  \rho_R(C) = \frac{[H_0 - H(C)]^+}{H_0}, \qquad H_0 > 0,
\]
with $\rho_R(C) = 0$ when $H_0 = 0$. The index $\rho_R(C) \in [0,1]$ equals 0 when context provides no reduction in uncertainty and 1 when it eliminates uncertainty entirely. The baseline $H_0$ is always estimable by withholding context; no oracle is required. In the experiments, $C = C^{(\alpha)}$ is parameterised by $\alpha$, so we write $\rho_R(\alpha)$; this is a property of the experimental design, not of $\rho_R$ itself.

The resolution ratio has a natural interpretation in terms of information theory. The numerator $H_0 - H(C)$ measures the reduction in response entropy attributable to context $C$: it estimates, at the instance level, how much uncertainty about the correct answer is resolved by the available information. This quantity is analogous to the pointwise information gain $I(Y; C) = H(Y) - H(Y \mid C)$, with $H_0$ playing the role of the marginal entropy $H(Y)$ and $H(C)$ the role of the conditional entropy $H(Y \mid C)$ evaluated at the observed context. Normalizing by $H_0$ yields a scale-free index that is comparable across questions with different baseline uncertainty. Unlike mutual information, $\rho_R$ is defined and estimable per instance from two sets of black-box samples, without distributional assumptions or access to model internals.

\subsection{Information Dependence Index}

The Information Dependence Index compares the target quantity $p(q,C)$ across evidence conditions, quantifying how predictive accuracy changes as context is removed.

We define the Information Dependence Index (IDI) as $\Delta = \mathrm{Acc}_1 - \mathrm{Acc}_0$. This measures the extent to which correct prediction depends on access to the context. The value $\Delta \approx 0$ means that the answer is recoverable without context, while $\Delta \approx 1$ implies that the answer depends entirely on the context.

Under the missing data perspective, $\Delta$ quantifies the loss in predictive accuracy induced by missing contextual information, analogous to comparing inference under complete and incomplete data in multiple imputation.

\subsection{Repeated Sampling as Implicit Multiple Imputation}

\begin{table}[t]
\centering
\caption{Conceptual correspondence between MI and LLM repeated sampling.}
\label{tab:mi_llm}
\begin{tabular}{ll}
\toprule
Multiple Imputation & LLM repeated sampling \\
\midrule
Estimand $Q$ (quantity of interest) & $y$ (correct answer to $q$) \\
Complete predictors $X = (X_\text{obs}, X_\text{mis})$ & Context $C = (C^{(\alpha)}, C_\text{mis})$ \\
Imputation model $p(X_\text{mis} \mid X_\text{obs})$ & Implicit: $p_\theta(C_\text{mis} \mid C^{(\alpha)}, q)$ \\
Analysis model $p(y \mid q, X_\text{obs}, X_\text{mis})$ & Answer model $p_\theta(y \mid C^{(\alpha)}, C_\text{mis}, q)$ \\
Completed-data estimates $\hat{Q}_l$ & Sampled answers $Y^{(l)}$ \\
Pooled point estimate $\bar{Q} = \frac{1}{m}\sum_l \hat{Q}_l$ & Empirical answer distribution $\{p_k\}$ \\
Within-imputation variance $\bar{U} = \frac{1}{m}\sum_l U_l$ & Token log-probability (not used) \\
Between variance $B = \frac{1}{m-1}\sum_l(\hat{Q}_l - \bar{Q})^2$ & $H = -\sum_k p_k \log p_k$ \\
Total variance $T = \bar{U} + (1+1/m)B$ & Not estimated (requires $\bar{U}$) \\
Proportion due to missingness $\lambda = B/T$ & Identifiable given $\bar{U}$ (not estimated here) \\
\bottomrule
\end{tabular}
\end{table}

Both MI and LLM are generative models that produce completions based on conditioning information. In LLMs the response is the final product presented to the user. In MI the generated data (imputations) are an intermediate step in a larger pipeline.

Table~\ref{tab:mi_llm} maps the MI framework onto LLM repeated sampling, with estimand $Q = y$, the correct answer to question $q$. The complete predictor set $X = (X_\text{obs}, X_\text{mis})$ corresponds to the full context $C = (C^{(\alpha)}, C_\text{mis})$. The imputation and analysis steps are fused inside the LLM: a single forward pass implicitly completes the missing context and produces a sampled answer $Y^{(l)}$, the completed-data estimate $\hat{Q}_l$, without an explicit intermediate representation of $C_\text{mis}$. When $y$ is numeric, Rubin's combining rules apply directly to the $\hat{Q}_l$. When $y$ is categorical text, between-imputation variance $B$ is undefined; Shannon entropy $H = -\sum_k p_k \log p_k$ over the counts of sampled answers serves as its natural generalization. In the categorical setting, entropy replaces variance as the dispersion functional. The natural analogue of within-imputation variance $U_l$ is the token log-probability of a single draw, a single-draw self-assessment of generation uncertainty available from some APIs. This paper focuses on between-draw variability as a function of context completeness and does not use token-level probabilities. Once $\bar{U}$ is estimated on a scale comparable to $B$, the proportion of variance due to missing information $\lambda = B / T$ becomes identifiable.

Confidence $c = \max_k p_k$ has no counterpart in the MI variance decomposition: it measures modal concentration rather than distributional spread, and can remain high even when entropy is large --- exactly the condition under which a proper imputer should signal high uncertainty.

\section{Experimental Setup}

\subsection{Dataset}

We use the SQuAD dataset \citep{rajpurkar2016} in its short-answer setting, where each answer is a span extracted from the associated context. The dataset consists of instances $(q, C, y)$, where $q$ is a question, $C$ a context paragraph, and $y$ the ground-truth short answer.

\subsection{Evidence Conditions}

We control the amount of available information through a parameter $\alpha \in [0,1]$, representing the fraction of the original context retained. For each instance $(q, C)$, we construct a truncated context $C^{(\alpha)}$ by retaining the first $100 \cdot \alpha$ percent characters of $C$. The extreme cases $\alpha = 1$ and $\alpha = 0$ correspond to the full and no-context conditions, respectively. Intermediate values of $\alpha$ induce graded levels of partial observability. Although this truncation strategy provides a simple and reproducible mechanism for degrading information, it does not guarantee uniform removal of answer-relevant content across instances.

\subsection{Response Generation}

For each input $(q, C)$, we generate $m$ stochastic responses $Y^{(1)}, \dots, Y^{(m)} \sim p_\theta(\cdot \mid q, C)$ using \texttt{gpt-4o-mini}, $m = 10$, and temperature-based decoding with $T=0.7$. This temperature setting introduces controlled variability while suppressing low-probability outputs \citep{kuhn2023}. Responses are constrained to a short-answer format (1--3 words) to reduce linguistic variability unrelated to epistemic uncertainty.

In practice, stochastic draws from a language model can be correlated, leading to repeated or similar outputs. To minimise such dependence, we obtain each sample via a separate API call, ensuring that the decoder is re-initialised for every draw. This is computationally costly; modern APIs support batched sampling (multiple completions per request) which would substantially reduce latency, at the cost of potentially higher within-batch correlation.

\subsection{Sampling Scheme}

In extractive QA datasets, multiple questions often share the same context, inducing dependence under row-level sampling. To avoid this, we sample at the context level.

Let the dataset consist of instances $(x_i, C_i, y_i)$ for $i = 1, \dots, n$, where identical contexts define a common group. Let $g = 1, \dots, G$ index distinct context groups. We sample $G' \le G$ context groups without replacement and select one instance uniformly per group.

We analyze the full sample and a subset capturing context dependence. For each instance, we compute $\Delta_i = \mathrm{acc}_{i,1} - \mathrm{acc}_{i,0}$ and $H_{i,0} = -\sum_k p_{ik} \log p_{ik}$. The \emph{context-sensitive (CS) subset} requires $\Delta_i \ge \tau_\Delta$ ($\tau_\Delta = 0.6$), $H_{i,0} \ge \tau_H$ ($\tau_H = 0.5$), and $H_{i,1} \le \tau_{H_1}$ ($\tau_{H_1} = 0.05$), ensuring that full context fully resolves the uncertainty.

All main results are reported on the full sample; the CS subset is used to isolate the regime where context is the dominant driver of response variability.

\subsection{Evaluation Metrics}

We evaluate model behavior using accuracy ($\mathrm{Acc}$), response entropy ($H$), sampling-based confidence ($c$), and the Resolution Ratio $\rho_R(\alpha)$. The primary analysis examines how accuracy behaves across levels of $\rho_R(\alpha)$: questions with high $\rho_R(\alpha)$ are those where context has resolved most baseline uncertainty, and accuracy should be correspondingly high if the uncertainty measures are well-calibrated to missing information.

\section{Results}

\subsection{Taxonomy of Model Behavior}

In the full sample, we distinguish three regimes of model behavior under context removal, characterized by accuracy and entropy in the no-context condition ($H_0$).

\paragraph{Memorized.}
$\mathrm{Acc}_1 = 1$, $\mathrm{Acc}_0 = 1$, and $H_0 \approx 0$.
The model returns the same correct answer regardless of context, reflecting parametric knowledge.

\paragraph{Biased.}
$\mathrm{Acc}_1 = 1$, $\mathrm{Acc}_0 = 0$, and $H_0 \approx 0$.
The model produces a consistent but incorrect answer without context, indicating systematic bias.

\paragraph{Uncertain.}
$\mathrm{Acc}_1 = 1$ and $H_0 > 0$.
The model produces diverse answers when context is removed, reflecting uncertainty induced by missing information.

Table~\ref{tab:taxonomy_examples} shows one representative instance per regime from the same context group, including sampled responses under full and no-context conditions and the corresponding accuracy, confidence and entropy estimates.

\begin{table*}[t]
\centering
\small
\begin{tabular}{p{0.09\linewidth} p{0.27\linewidth} p{0.28\linewidth} p{0.18\linewidth}}
\toprule
\textbf{Regime} & \textbf{Full context responses} & \textbf{No context responses} & \textbf{Estimates} \\
\midrule

\multicolumn{4}{l}{\textbf{Q:} Which NFL team represented the AFC at Super Bowl 50? \quad \textbf{Truth:} Denver Broncos} \\[2pt]
\textbf{Memorized}
&
\begin{tabular}[t]{@{}l@{}}
Denver Broncos \\
Denver Broncos \\
Denver Broncos \\
Denver Broncos \\
Denver Broncos
\end{tabular}
&
\begin{tabular}[t]{@{}l@{}}
Denver Broncos \\
Denver Broncos \\
Denver Broncos \\
Denver Broncos \\
Denver Broncos
\end{tabular}
&
\begin{tabular}[t]{@{}l@{}}
$\mathrm{Acc}_1 = 1.0$ \\
$\mathrm{Acc}_0 = 1.0$ \\
$\Delta = 0.0$ \\
$c_1 = 1.0$, $c_0 = 1.0$ \\
$H_1 = 0$, $H_0 = 0$
\end{tabular}
\\
\midrule

\multicolumn{4}{l}{\textbf{Q:} What does AFC stand for? \quad \textbf{Truth:} American Football Conference} \\[2pt]
\textbf{Biased}
&
\begin{tabular}[t]{@{}l@{}}
American Football Conference \\
American Football Conference \\
American Football Conference \\
American Football Conference \\
American Football Conference
\end{tabular}
&
\begin{tabular}[t]{@{}l@{}}
Asian Football Confederation \\
Asian Football Confederation \\
Asian Football Confederation \\
Asian Football Confederation \\
Asian Football Confederation
\end{tabular}
&
\begin{tabular}[t]{@{}l@{}}
$\mathrm{Acc}_1 = 1.0$ \\
$\mathrm{Acc}_0 = 0.0$ \\
$\Delta = 1.0$ \\
$c_1 = 1.0$, $c_0 = 1.0$ \\
$H_1 = 0$, $H_0 = 0$
\end{tabular}
\\
\midrule

\multicolumn{4}{l}{\textbf{Q:} What was the theme of Super Bowl 50? \quad \textbf{Truth:} Golden anniversary} \\[2pt]
\textbf{Uncertain}
&
\begin{tabular}[t]{@{}l@{}}
Golden anniversary \\
Golden anniversary \\
Golden anniversary \\
Golden anniversary \\
Golden anniversary
\end{tabular}
&
\begin{tabular}[t]{@{}l@{}}
Celebration of greatness \\
Legacy Game \\
Celebration of Champions \\
Celebration of Champions \\
Broncos vs.\ Panthers
\end{tabular}
&
\begin{tabular}[t]{@{}l@{}}
$\mathrm{Acc}_1 = 1.0$ \\
$\mathrm{Acc}_0 = 0.0$ \\
$\Delta = 1.0$ \\
$c_1 = 1.0$, $c_0 = 0.4$ \\
$H_1 = 0$, $H_0 \approx 1.33$
\end{tabular}
\\

\bottomrule
\end{tabular}
\caption{Illustrative examples of the three regimes. The memorized regime reflects context-independent recall ($\Delta = 0$, $H_0 = 0$). The biased regime shows deterministic but incorrect responses under missing context ($\Delta = 1$, $H_0 = 0$). The uncertain regime exhibits dispersion in responses when context is removed ($H_0 > 0$), indicating uncertainty induced by missing information. Only the uncertain regime exhibits non-zero entropy under missing context.
}
\label{tab:taxonomy_examples}
\end{table*}

In a sample of $N = 400$ contexts, the distribution of regimes is 54 memorized (13.5\%), 41 biased (10.2\%), 175 uncertain (43.8\%), and 130 other (32.5\%). The uncertain regime is the largest, reflecting that most questions cannot be answered without context but admit diverse responses when it is missing. The context-sensitive (CS) subset ($N = 148$) retains only cases where full context completely resolves uncertainty, so that $\rho_R(1) = 1$ for all members by construction.

\subsection{Context Truncation Effects}

Figure~1 shows accuracy, confidence $c$, and entropy $H$ as functions of $\alpha$ for the full sample and the CS subset. In both samples, accuracy decreases monotonically as context is removed. Confidence also declines, but much more slowly, remaining relatively high even as accuracy collapses—most notably in the CS subset, where accuracy falls to 0.05 while confidence remains at 0.52. In contrast, entropy closely mirrors the loss in accuracy.

\begin{figure}[tbh]
\centering
\includegraphics[width=1.0
\linewidth]{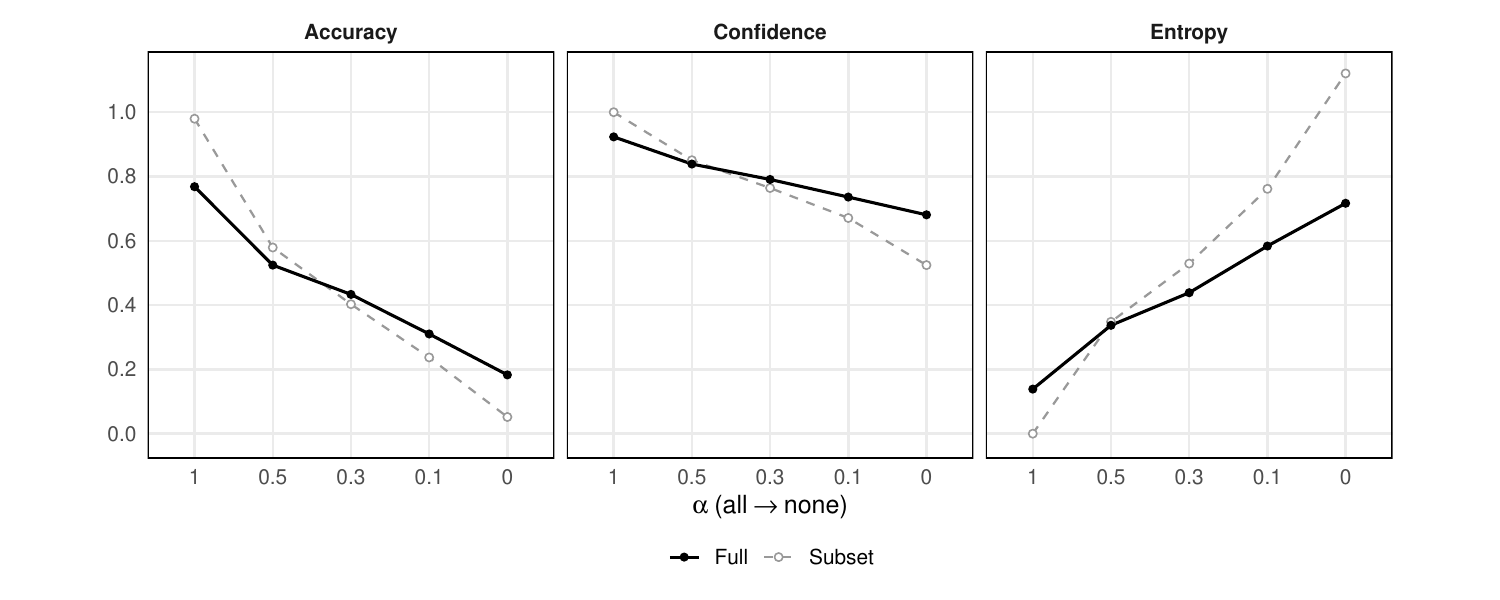}
\caption{Accuracy, confidence, and entropy as a function of context completeness.}
\end{figure}

Figure~2 shows the effect of decreasing context availability on uncertainty, calibration, and overconfidence. In the full sample, the resolution ratio $\rho_R$ indicates that full context resolves approximately 80\% of the baseline uncertainty, with progressively smaller reductions as context is truncated. The calibration plot reveals a substantial mismatch between observed accuracy and confidence $c$, with overconfidence increasing as context is removed.

\begin{figure}[tbh]
\centering
\includegraphics[width=1.0\linewidth]{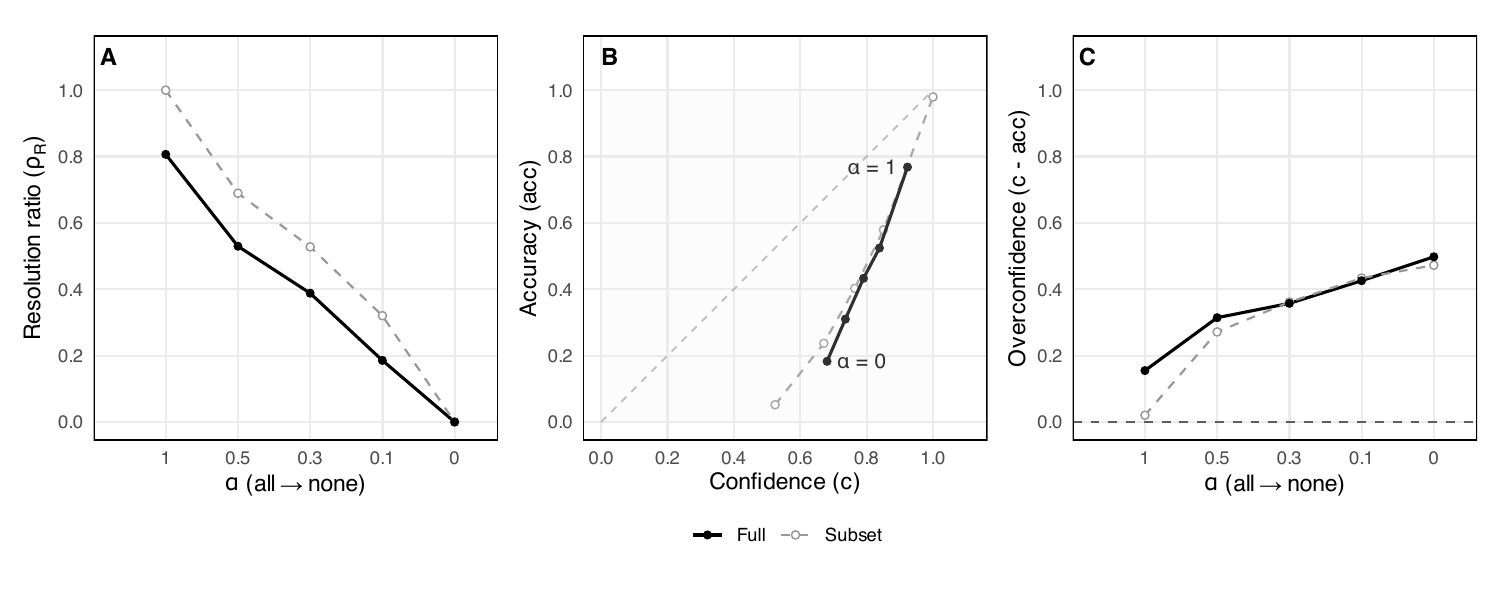}
\caption{
(A) Fraction of baseline uncertainty resolved by context ($\rho_R$).
(B) Calibration plot showing accuracy versus confidence; the diagonal indicates perfect calibration.
(C) Overconfidence ($c - \mathrm{acc}$)}
\end{figure}

\subsection{Uncertainty and Predictive Accuracy}

\begin{figure}[tbh]
\centering
\includegraphics[width=1.0\linewidth]{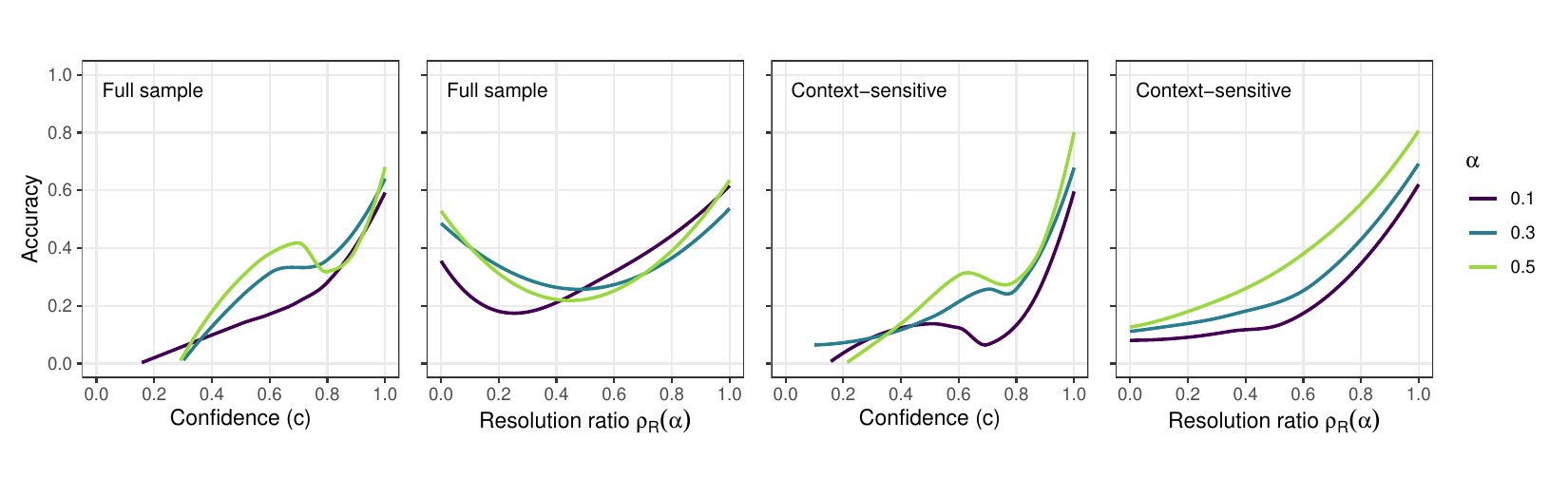}
\caption{Accuracy as a function of confidence and resolution ratio $\rho_R(\alpha)$, for the full sample and the context-sensitive (CS) subset, at $\alpha \in \{0.1, 0.3, 0.5\}$. Curves are loess smoothers.}
\label{fig:gamma_scatter}
\end{figure}

Figure~\ref{fig:gamma_scatter} shows that $\rho_R(\alpha)$ predicts accuracy monotonically in the CS subset, with curves ordered by $\alpha$. Questions for which context resolves more uncertainty are answered more accurately, and greater resolution consistently translates into higher accuracy at each evidence level. Confidence does not exhibit this ordering: it shows a non-monotone pattern in both samples, driven by the biased regime, where high confidence is associated with consistently incorrect answers.

Table~\ref{tab:r2} quantifies the advantage of entropy over confidence as a predictor of accuracy. Entropy consistently explains more variance across all conditions, with the gap largest at $\alpha = 0.5$ in the full sample ($R^2 = 0.243$ vs.\ $0.186$), where entropy explains approximately 31\% more variance than confidence. The advantage persists in the CS subset, where both predictors achieve higher absolute $R^2$ due to the removal of memorized and biased questions.

\begin{table}[h]
\centering
\caption{$R^2$ from quadratic regression of accuracy on confidence ($c$) and entropy ($H$), by sample and context level $\alpha$. Quadratic terms improve fit over linear by 2--6 percentage points; cubic terms add negligibly.}
\label{tab:r2}
\begin{tabular}{llccc}
\toprule
Sample & $\alpha$ & $R^2(c)$ & $R^2(H)$ & $\Delta R^2$ \\
\midrule
Full & 0.1 & 0.212 & 0.249 & 0.037 \\
Full & 0.3 & 0.202 & 0.240 & 0.037 \\
Full & 0.5 & 0.186 & 0.243 & 0.057 \\
\midrule
CS   & 0.1 & 0.251 & 0.306 & 0.055 \\
CS   & 0.3 & 0.289 & 0.335 & 0.046 \\
CS   & 0.5 & 0.307 & 0.355 & 0.048 \\
\bottomrule
\end{tabular}
\end{table}

\section{Discussion}

\subsection{The MI framing as an evaluation standard.}

Framing repeated LLM sampling as implicit multiple imputation provides a normative criterion for uncertainty. A proper imputation model propagates uncertainty due to missing data to downstream predictions; an improper model suppresses it. Accordingly, predictive uncertainty should increase as information is removed.

The distinction has practical consequences. Deterministic imputation (e.g., replacing missing BMI by conditional means) reduces variability and underestimates quantities that depend on the distribution, such as obesity prevalence \citep[Section~9.3]{vanbuuren2018}. The same mechanism is at work here: suppressing variability induced by missing information leads to overconfident and potentially misleading conclusions. Standard point-prediction criteria such as mean squared error reinforce this behavior, as they reward accurate reconstruction and produce point estimates that do not propagate uncertainty due to missing information.

MI also suggests concrete extensions beyond answer selection. For example, an LLM could be prompted to analyze a caregiver report and produce a developmental score under incomplete information. Repeated LLM sampling would then yield multiple imputations of the latent score, enabling estimation of a prediction interval via standard MI pooling rules. This illustrates how LLMs can be used not only to generate point predictions, but to quantify uncertainty in downstream quantities of interest.

The MI analogy also suggests directions for improving LLM uncertainty propagation, not just evaluating it. A plausible source of overconfidence under missing context is the softmax operation: by normalizing over the full output distribution at each token, softmax concentrates probability mass on the modal response and suppresses variability, even when the input is uninformative. This produces the improper imputer behavior observed in the biased regime: near-zero entropy despite high missingness. Predictive mean matching (PMM) offers a principled alternative \citep{vanbuuren2018}: rather than passing through softmax, the model draws from an implicit donor pool encoded in its parametric knowledge. Asked about eye color with no context, it samples from the marginal population distribution; given hair color, the effective donor pool narrows and entropy falls accordingly. Repeated sampling then yields a distribution whose spread directly reflects residual uncertainty. This reframes the problem: softmax may not be optimal under partial observability, and MI theory suggests that donor-based sampling should produce better-calibrated uncertainty in this regime.

\subsection{Entropy as a responsive uncertainty measure.}

Four empirical results support entropy as a suitable black-box uncertainty measure under incomplete context. First, entropy increases monotonically as context is removed, closely tracking the loss in accuracy, while confidence remains high even as predictive performance collapses. Second, full context resolves approximately 80\% of baseline entropy in the full sample ($\rho_R(1) = 0.81$), confirming that entropy responds strongly to the presence of informative context. Third, a quadratic regression of accuracy on entropy yields consistently higher $R^2$ than the same regression on confidence across all evidence levels and both samples; the gap reaches 0.057 at $\alpha = 0.5$ in the full sample, where entropy explains approximately 31\% more variance in accuracy. Fourth, the resolution ratio $\rho_R(\alpha)$ predicts accuracy monotonely and is ordered by $\alpha$ in the CS subset, whereas confidence shows a non-monotone pattern driven by the biased regime. Together, these results indicate that entropy behaves consistently with the MI criterion for proper uncertainty propagation, while confidence does not.

The sampling-based approach underlying these results is broadly applicable. Estimating entropy requires only repeated draws from the model at standard temperature: no access to weights, logits, or internal activations, and no entailment classifier. It is model-agnostic, requires no task-specific adaptation, and can be applied to any LLM accessible via an API. The diagnostic $\rho_R(\alpha)$ is equally general: it requires only two sets of samples, with and without context, and produces an interpretable index of how much context contributes to resolving uncertainty. Both quantities are thus deployable in practice wherever repeated sampling is feasible.

\subsection{Limitations.}
First, entropy is computed over normalized answer strings, so semantically equivalent paraphrases are treated as distinct responses. As a result, entropy may overstate dispersion for verbose or paraphrase-prone models. Second, the experiments use SQuAD short factual answers, where correctness is binary. Extending the analysis to longer-form generation tasks, where correctness is graded or multi-dimensional, would require adapted definitions of both entropy and accuracy. Third, we study only one missing data mechanism: prefix truncation. This mechanism is conservative relative to MCAR, since contiguous removal is more likely to eliminate answer-relevant content than random masking. Whether entropy responds as faithfully under other missingness mechanisms remains an open question, though the directional prediction is clear. Fourth, the context-sensitive (CS) subset is defined partly in terms of entropy ($H_0 \geq 0.5$, $H_1 \leq 0.05$), so its favorable $\rho_R$ pattern is partly by construction. The full-sample results, which impose no entropy criterion, show the same qualitative ordering, though diluted by memorized and biased questions.

\section{Conclusion}

Multiple imputation provides a mature framework for reasoning about prediction under incomplete data, with well-defined criteria for proper uncertainty propagation. Importing these criteria into the LLM setting yields a practically useful, theory-inspired standard for evaluating uncertainty measures, independent of any particular task or model.

The between-imputation variance analogue (entropy as an approximate counterpart to $B$), the uncertainty-resolved diagnostic $\rho_R(\alpha)$, and the proper/improper imputation distinction are directly portable. Extensions to missing-data mechanisms (e.g., MCAR, MAR, MNAR), not studied here, offer natural directions for future work.

This connection may prove productive beyond the specific setting studied here, opening a channel between the statistical missing-data literature and the rapidly growing literature on LLM uncertainty.

\newpage

\bibliographystyle{plainnat}
\bibliography{refs}

\newpage

\end{document}